# A Spiking Binary Neuron – Detector of Causal Links

**Mikhail Kiselev, Denis Larionov, Andrey Urusov**
Chuvash State University
Cheboksary, Russia
`mkiselev@chuvsu.ru`

August 17, 2023

## Abstract

Causal relationship recognition is a fundamental operation in neural networks aimed at learning behavior, action planning, and inferring external world dynamics. This operation is particularly crucial for reinforcement learning (RL). In the context of spiking neural networks (SNNs), events are represented as spikes emitted by network neurons or input nodes. Detecting causal relationships within these events is essential for effective RL implementation. This research paper presents a novel approach to realize causal relationship recognition using a simple spiking binary neuron. The proposed method leverages specially designed synaptic plasticity rules, which are both straightforward and efficient. Notably, our approach accounts for the temporal aspects of detected causal links and accommodates the representation of spiking signals as single spikes or tight spike sequences (bursts), as observed in biological brains. Furthermore, this study places a strong emphasis on the hardware-friendliness of the proposed models, ensuring their efficient implementation on modern and future neuroprocessors. Being compared with precise machine learning techniques, such as decision tree algorithms and convolutional neural networks, our neuron demonstrates satisfactory accuracy despite its simplicity. In conclusion, we introduce a multi-neuron structure capable of operating in more complex environments with enhanced accuracy, making it a promising candidate for the advancement of RL applications in SNNs.

## 1 Introduction

If we aim to develop an intelligent system based on neural networks capable of learning adaptive behavior towards specific objectives, it is imperative to equip it with the capacity to identify and incorporate causal relationships between events occurring both within the network and in the external environment. These relationships may encompass consistent sequences of patterns forming a unified spatiotemporal pattern, directives to actuators generated by the network, and subsequent responses from the external environment, as well as events preceding a reward and the reward itself. Therefore, the capability to discern causes and effects should constitute a foundational functionality within network structures or individual neurons. In most scenarios, the learning network lacks access to a priori knowledge outlining causal relationships in its environment. Instead, it must deduce them from observed temporal alignments of different events, under the assumption that if event B is frequently witnessed within a specific time frame following event A, then A serves as the cause and B as the consequence.

In this research, we investigate the integration of this particular functionality within spiking neural networks (SNNs), specifically within an individual neuron embedded in the SNN. In the context of SNN, information is encoded through sequences of spikes, making it necessary to frame this problem within the spike domain.

We shall formally define the task to be addressed by this neuron, referred to as the causal relationship detector. The neuron receives spikes from a group of other neurons. It is postulated that the concurrent activation of an unidentified subset of these presynaptic neurons signifies event A (the cause), while the activation of a distinct neuron indicates event B (the consequence). Given the assumption that event B consistently or frequently occurs within a time window of $T_P$ following event A, the neuron's objective is to identify the specific presynaptic neurons belonging to this subset and fire whenever event A occurs. Similar to other learning challenges encountered in SNNs, this problem is resolved by introducing appropriate adjustments to the synaptic weights of the neuron-detector. Importantly, and notably complicating the matter, both the presynaptic neurons and the neuron-detector itself may produce not only individual isolated spikes but also tightly clustered and prolonged spike sequences. This aspect introduces a significant challenge, as it limits the applicability of the majority of existing synaptic plasticity models that rely on the relative timing of individual pre- and post-synaptic spikes.

Numerous studies have demonstrated how SNN can express causal relationships between observations. However, the present research offers a unique combination of three distinctive attributes:

1. Causal relationships are identified by a single neuron, rather than relying on a network-wide approach.
2. The temporal dimension of causal relationships is emphasized, with events-causes occurring certain time before their corresponding consequences.
3. Specially designed local synaptic plasticity rules are used for learning.

The majority of works consider this problem without a temporal aspect. In this context, the problem is often called Bayesian inference or causal inference. It has a close relation to supervised learning, where the network should determine which factors, either independently or in combination, most reliably indicate that the given object belongs to the target class. Recent examples of such research taking various approaches to tackle causal/Bayesian inference problems can be found in [1, 2, 3]. It is noteworthy that one of the primary objectives of these works is supervised learning, which does not explicitly incorporate time. Even when dealing with time series data, the mentioned works treat the entire time series as a single entity, assigning a label to the entire sequence. In contrast, our research aligns more closely with reinforcement learning where all the signals (input signals, network commands, and reward/punishment) exist within the continuous time where time delays carry significance.

Furthermore, our approach resonates with Friston's concepts of free energy [4], as understanding the "expected events" is essential for quantifying the "surprise" or unexpected events in terms of free energy. In essence, this research delves into the intricacies of recognizing causal relationships within the temporal context, which holds particular relevance for dynamic, time-dependent systems.

It's worth noting that our research problem bears resemblance to another significant area of interest among machine learning and neural network researchers, namely, time series forecasting. Time series forecasting techniques aim to predict the future values of specific variables (whether discrete or continuous) based on their current and recent values, and possibly those of other related variables. Naturally, if we can identify causal relationships between the values of certain variables or the states of an object and the values of specific parameters in the future, it equips us with a tool for predicting these future values. However, our primary objective differs from traditional time series forecasting, as we are not focused on predicting the exact value of a particular variable at an exact point in time. Instead, our aim is to infer causal rules that indicate that after event A, it is expected that event B will occur within a time interval of length $T_P$. Consequently, our problem can be more aptly characterized as future event prediction.

Remarkably, there are relatively few applications of SNNs to this problem. One of such approaches is described in [7]. It utilizes the NeuCube system [5], which is centered around the Liquid State Machine (LSM) [6]. The LSM is a large, chaotic, non-plastic SNN designed to transform a combination of time series data and static (or slowly changing) parameters into a high-dimensional representation, represented by the current firing frequencies of neurons within the LSM. Due to the great number of neurons in the LSM, representations of various spatiotemporal patterns in the form of LSM neuronal activity tend to be linearly separable. Consequently, classification problems related to such representations can be efficiently solved using simple linear classifiers. As described in [7], several instances of NeuCube's application to rare event forecasting have been demonstrated. One specific example, the prediction of strokes, is examined in greater depth in [8]. While the LSM-based approach has demonstrated success across a wide range of tasks, it has a notable drawback in that it requires a large, computationally intensive LSM to achieve efficiency. In contrast, our approach effectively addresses a similar problem using just a single neuron, offering a more resource-efficient solution.

Paper [9] shows how special SNN structures can be used to infer the graph of causal relationships, but, again, without temporal aspect as mentioned before.

At last, there exists another branch of SNN research closely aligned with our study. As described below, we employ a combination of two synaptic plasticity models—commonly referred to, albeit in a very approximate sense, as Hebbian plasticity and dopamine plasticity—to address the problem of identifying causal links. Hebbian plasticity, when applied to the plasticity of spiking neurons, is often denoted as the STDP plasticity model (Spike Timing Dependent Plasticity) [10], while dopamine plasticity is typically associated with reward-related effects. The collective terminology for these merged plasticity models is R-STDP (Reward Spike Timing Dependent Plasticity). Numerous works have explored and proposed various R-STDP models [11, 12, 13, 14], with some of them having already undergone testing in real-world applications [15]. Presently, there is no universally accepted consensus on how to best combine these two types of synaptic plasticity, resulting in a wide array of proposed models. Moreover, unlike our approach where reward takes the form of a spike signal, these models typically represent reward as a global real-valued variable. To the best of our knowledge, no prior work has employed similar synaptic plasticity rules for the specific purpose of detecting causal links.

Furthermore, our specific objective was to craft a simplified plasticity model to facilitate efficient implementation on contemporary and forthcoming neuroprocessors. According to work [16], there is a clear trend in the emergence and development of software and hardware systems that allow not only the inference of convolutional neural networks converted into SNN form, but also the use of arbitrary models of spiking neurons and plasticity rules, opening up the possibility of continuous learning.

In the subsequent section, we will provide a detailed description of our innovative synaptic plasticity model, which combines Hebbian (effectively, anti-Hebbian) and dopamine plasticity. Following this, we will consider the application of this model to the problem of reward prediction in reinforcement learning (RL), using as an example the task of reward prediction in the "ping-pong" RL task. In conclusion, we will outline our vision of how neurons of this kind can form network structures capable of inferring complex graphs of causal relationships essential for constructing external world models in RL. Subsequently, we will assess the advantages and limitations of our approach and outline future research plans in this direction.

## 2 Materials and Methods

In our research, we investigate the learning process of a single spiking neuron in the context of identifying causal relationships among events. This neuron is connected to a group of presynaptic neurons denoted as set $C$, whose activity is presumed to represent various events in a general sense. In our study, we interpret these events as potential triggers for another event, which we will refer to as the "target event". This target event is signaled by spikes from a separate presynaptic neuron denoted as $S$, which is not a part of set $C$. The time moment of $j$-th firing of the $i$-th presynaptic neuron from the set $C$ is $t_{ij}$. The moments when the neuron $S$ fires will be denoted as $T_i$. We say that some event is a cause of the target event if the target event is frequently observed not later than the time $T_P$ after that event. As we have mentioned, these possible causes of the target event are indicated by the specific activity of the presynaptic neurons, which the learning neuron should try to recognize. $T_P$ is the temporal constant fixing time scale of the concrete domain or task. It is assumed that the target events are rare – it means that $T_P$ is much less than the minimum value of interspike intervals $T_i$ - $T_{i-1}$. It is the only important assumption – without it, our problem of finding causal relationships seems to become senseless.

We introduce the concept of a "target period", which encompasses the time interval of length $T_P$ preceding each $T_i$. The learning neuron should mark the target periods by its activity (spikes emitted by it at the times $T_i^*$). If it learns to do it with sufficient accuracy, it signifies that it has successfully recognized the causal link between a specific event (corresponding to the distinct activity of a presynaptic neuron that triggers the neuron to fire) and the target event. To assess the accuracy of this recognition, we introduce the concept of a "prediction period". Each prediction period begins at some moment $T_i^*$ and ends either after time $T_P$ from that moment or at some moment $T_i$ – whichever occurs first. The total duration of time $t_{err}$, during which the target period and prediction period do not overlap, serves as a natural metric for measuring the inaccuracy in predicting the target event. The learning neuron's objective is to maximize the metric represented by the equation:

$$R = 1 - t_{err} / t_{tar} \qquad (1)$$

where $t_{tar}$ signifies the total duration of target periods.

In our study, we use the simplest neuron model called "binary neuron". This neuron operates within discrete time intervals. Every time interval, it receives spikes via its plastic synapses with the weights $w_i$. If the sum of weights of synapses receiving spikes in this time interval is greater than the threshold value $H$, the neuron fires. This choice makes our result general – in fact, it does not depend on the concrete neuron model. After the respective time discretization, any neuron model can be approximated by the binary neuron which retains the main spiking neuron property – neuron fires when several strong excitatory synapses obtain spikes during short time period. In order to make the weights $w_i$ dimensionless, we set $H = 1$.

## 2.1 General Idea of the Method and the Synaptic Plasticity Rules Used

We assume that information about potential causal events leading to the target event is encoded within spikes originating from the presynaptic neurons in set $C$. The synapses responsible for transmitting these spikes are plastic, and their synaptic weights should be modified in such a way that would force the neuron to fire during the target period.

Activity of the learning neuron and modification of its synaptic weights should be interrelated in the following way:

A. Untrained neuron should be inactive – the plasticity mechanism should strengthen those synapses which would force the neurons to fire in the correct time. For this reason, we set weights of all plastic synapses to zero at the learning start.
B. If the neuron fires in the wrong time (outside of target periods) then the synapses which helped it to fire should be depressed.
C. If the neuron fires in the correct time then nothing should be done with its synapses – otherwise modifying its synaptic weights can move it from a trained state.

These behaviors are achieved through the specific design of synaptic plasticity rules. It is important that properties of the plastic synapses are totally different from the single synapse through which the presynaptic neuron $S$ is connected. We will call it the "dopamine" synapse" because spikes coming to it control the plasticity of all other synapses.

The principles A, B, and C specified above are fulfilled due to the following combination of the two plasticity rules:

1. Dopamine plasticity. Every time the neuron receives a spike from the $S$ neuron all the plastic synapses having obtained a spike during the time $T_P$ before this "dopamine" spike are potentiated.
2. Anti-Hebbian plasticity. All synapses contributing to neuron firing are depressed.

It is evident that the equilibrium between dopamine plasticity and anti-Hebbian plasticity, when properly balanced, satisfies the conditions A to C, ensuring the successful training and functioning of the neuron.

## 2.2 Synaptic Plasticity Model in Detail

Similar to our previous research works [21, 22], the synaptic plasticity rules are additive and applied to a variable termed "synaptic resource," denoted as $W$, rather than directly to the synaptic weight, denoted as $w$. There is functional dependence between $W$ and $w$ expressed by the formula:

$$w = w_{\min} + \frac{(w_{\max}-w_{\min})\max(W,0)}{w_{\max}-w_{\min}+\max(W,0)} \qquad (2)$$

where $w_{min}$ and $w_{max}$ are constants. It is obvious that $w$ values lay inside the range [$w_{min}$, $w_{max}$) - while $W$ runs from -∞ to +∞. In this study, $w_{min} < 0$, while $w_{max} > 0$ so that synaptic plasticity can make excitatory synapse inhibitory and vice versa.

As previously mentioned, the synaptic plasticity model comprises two distinct and independent components, which are elaborated upon in subsections 2.2.1 and 2.2.2.

### 2.2.1 Anti-Hebbian Plasticity

The standard STDP (spike timing dependent plasticity) model [10] states that spikes coming short time before postsynaptic spike emission potentiate the synapses receiving them. This concept aligns with Donald Hebb's principle, which asserts that synaptic plasticity should encode causal relationships within neuron firings; in essence, synapses responsible for inducing neuron firing should be strengthened. This principle has been proven by plenty of neurophysiological observations. However, in-depth investigations into plasticity within biological neurons have revealed multiple instances of entirely distinct synaptic plasticity models existing in nature [17, 18]. Furthermore, examples of plasticity rules acting in the direction opposite to Hebbian principle (anti-Hebbian plasticity) have been observed in different organisms [19]. It makes us conclude that different kinds of synaptic plasticity are suitable for the solution of different problems. Moreover, the standard STDP model becomes senseless or self-contradictory in the case (which is quite common in the biological brain) when presynaptic and postsynaptic spikes are not stand alone in time but are packed in tight sequences (spike trains). In this case, it is senseless to say that presynaptic spike comes before or after postsynaptic spike because there are many postsynaptic spikes in the close neighborhood before and after the given presynaptic spike.

For these reasons, we have devised our own variant of the anti-Hebbian plasticity model and applied it to the problem at hand. Let us elaborate on this model.

As previously mentioned, weight modifications in the standard STDP are bound to single pre- and post-synaptic spikes. However, in the presence of spike trains, these rules lose their applicability. Consequently, in our model, the synaptic plasticity acts are bound to postsynaptic spike trains instead of single spikes. We refer to these spike trains as "tight spike sequences" (TSS). Specifically, taking the constant $ISI_{max}$ as a measure of "tightness" of TSS, we define TSS as a sequence of spikes adhering to the following criteria:

1. There were no spikes during time $ISI_{max}$ before the first spike in TSS;
2. Interspike intervals for all neighboring spikes in TSS are not greater than $ISI_{max}$;
3. There are no spikes during time $ISI_{max}$ after the last spike in TSS.

In this work, $ISI_{max}$ is set to the value of $T_P$.

Our anti-Hebbian plasticity model adheres to the following rules:

1. Resource of any synapse can change at most once during a single TSS. Here and below, TSS refers to postsynaptic spike train.
2. Resources of only those synapses are changed which receive at least one spike during TSS.

All synaptic resources are changed (decreased) by the same value $d_H$ independently of exact timing of presynaptic spikes.

## 2.2.2 Dopamine Plasticity

The described neuron has a plasticity-inducing synapse (from *S*). When it obtains a spike, synaptic resources of all plastic synapses of the neuron having received at least one presynaptic spike during the time interval of the length $T_P$ in the past are modified (increased) by the same value $d_D$.

## 2.2.3 Neuron Stability

In our model, the synaptic plasticity values $d_H$ and $d_D$ are not constant. Instead, they are dynamic and exhibit characteristics that adapt over learning. Initially, during the early stages of learning, these values need to be sufficiently large. However, for a well-trained neuron that consistently makes accurate predictions, they should approach zero. This adaptation is crucial to prevent further modifications to the neuron's synaptic weights, which could potentially disrupt its established trained state. To account for this adaptive behavior, we introduce an additional component into the neuron's state, denoted as "stability."

The synaptic plasticity values decrease exponentially to zero as the stability value grows, governed by the following expressions:

$$d_H = \overline{d_H}\min(2^{-s}, 1), \quad d_D = \overline{d_D}\min(2^{-s}, 1). \tag{3}$$

Here, $\overline{d_H}$ and $\overline{d_D}$ are constants of the neuron model. In order to balance anti-Hebbian and dopamine plasticity (that is necessary for holding the condition C from subsection 2.1) we set $\overline{d_H} = \overline{d_D}$.

The neuron stability value changes in the two situations:

1. It is decreased by the constant $d_s$ every TSS.
2. It is adjusted by the value $d_s\max\left(2 - \frac{|t_{TSS} - ISI_{max}|}{ISI_{max}}, -1\right)$ whenever a presynaptic dopamine spike is received. Here $t_{TSS}$ is the time interval between the most recent TSS onset and the dopamine spike.

We see that if TSS began exactly the time $ISI_{max}$ (= $T_P$) ago before dopamine spike (i.e. target event) the neuron stability increase is maximum and is equal to $d_s$ – if to take into account its decrease by $d_s$ in accordance with rule 1. This corresponds to the most accurate prediction of the target event and serves as evidence that the neuron is well trained. Conversely, if a dopamine spike occurs when the neuron has remained inactive for an extended period, it suggests inadequate training, and as a result, its stability is decreased by $d_s$ to facilitate further training.

## 2.3 The Test Task – Find the Cause of Obtaining Reward in the Ping-Pong ATARI Game

The described technique holds great promise in the realm of reinforcement learning (RL). While supervised learning can be conceptualized as identifying causal relationships between predictors as causes and the target value as a consequence, RL tasks encompass a broader spectrum of causal link determination, explicitly incorporating the element of time. Reward signals come rarely and, possibly, with significant delay from the world states or the agent actions they evaluate. In order to overcome problems with insufficient evaluation signal frequency, the mechanism of intermediate goals should be utilized, but it is also based on inferring causal links.

Futhermore, the key point of the most comprehensive realization of RL, known as model-based RL, is creation by the agent of the internal model of world dynamics and world reactions to agent actions. The model creation mechanism unavoidably includes inference of a network of causal relationships between world state changes and agent actions. Thus, it is entirely reasonable to assert that causal link inference is the most basic operation in RL.

Keeping this in mind, we selected the first task to test capability of the neuron model proposed from the RL domain. Namely, we chose one of the RL tasks drawn from the well-established ATARI games benchmark set [20]. This task centers around the computerized ping-pong game, where a ball traverses within a square area, rebounding off its walls. The area has only three walls. Instead of the left wall, the racket moves in the vertical direction on the left border of this square area. The racket is controlled by the agent which can move it up and down. When the ball hits the racket it bounces back and the agent obtains a reward signal. If the ball crosses the left border without hitting the racket the agent gets punishment and the ball is returned to a random point of the middle vertical line of the area, gets random movement direction and speed and the game continues. Using the reward/punishment signals received, the agent should understand that its aim is to reflect the ball and learn how to do it.

In our example, the network (in fact, a single neuron) should solve the very first problem – to "understand" what conditions cause obtaining reward in the near future.

Let us describe the input information coming to the neuron's plastic synapses. This information includes the current positions of the ball and the racket and the ball velocity. While the ultimate formulation of this problem would involve primary raster information (i.e., the screen image), our focus here extends beyond computer vision and delves into the realm of causal relationships. Consequently, we assume that preceding layers have already processed the primary raster data and converted it into the spike-based description of the world state, which forms the basis of our neuron's input.

The input nodes that are sources of spikes sent to the learning neuron are subdivided into the following sections:

1. The ball X coordinate Consists of 30 nodes capturing the ball's horizontal position. The horizontal dimension is broken to 30 bins. When the ball is in the bin *i*, the *i*-th node emits spikes with frequency 300 Hz. To establish spatial and temporal scales we assume that the size of the square area is 10×10 cm (so that the boundary coordinates are ±5 cm) and the discrete emulation time step is 1 msec.
2. The ball Y coordinate. Consists of 30 nodes capturing the ball's vertical position. Similar to X but for the vertical axis.
3. The ball velocity X component. Consists of 9 nodes capturing the ball's horizontal velocity. When the ball is reset in the middle of the square area, its velocity is set to the random value from the range [10, 33.3] cm/sec. Its original movement direction is also random but it is selected so that its X component would not be less than 10 cm/sec. The whole range of possible ball velocity X component values is broken to 9 bins such that the probabilities to find the ball at a random time moment in each of these bins are approximately equal. While the ball X velocity is in some bin, the respective input node emits spikes with 300 Hz frequency.
4. The ball velocity Y component. Consists of 9 nodes capturing the ball's vertical velocity. The same logic as for the X velocity component.
5. The racket Y coordinate. Consists of 30 nodes capturing the racket's vertical position. Similar to the ball Y coordinate. The racket size is 1.8 cm so that the racket takes slightly more than 5 vertical bins.
6. The relative position of the ball and the racket in the close zone. Consists of 25 nodes capturing the ball's positions close to the racket. The square visual field of size 3×3 cm moves together with the racket so that the racket center is always at the center of the left border of this visual field. The visual field is broken to 5×5 square zones. When the ball is in some zone, the respective input node fires with frequency 300 Hz.

In total, there are 133 input nodes transmitting their spikes to the learning neuron. The neuron's objective is to discern the conditions that lead to obtaining a reward signal within a 50 msec timeframe, and accordingly, we set $T_P = 100$ msec.

## 2.4 Selection of the Neuron Parameters Using the Genetic Algorithm

While our neuron model appears to be relatively simple, it incorporates several parameters that require appropriate calibration. There are four key parameters in our model:

- The maximum synaptic resource change $\overline{d_H}$. This parameter controls learning speed. Its low values make learning slow, high values may make it unstable.
- The minimum synaptic weight value $w_{min}$. It is negative.
- The maximum synaptic weight value $w_{min}$.
- The stability change speed $d_s$.

The optimum values of $\overline{d_H}$ and $d_s$ are determined by strength of causal links – in case of weak determinism or high noise, great values of these parameters will lead to learning instability. $w_{min}$ and $w_{max}$ should be selected on the basis of mean input spike flow – the input node count and the mean spike frequency per node.

Although the general principles for setting these parameters are sufficiently clear, we decided to find their optimum values using the genetic algorithm and the very wide search ranges: [0.03, 1] for $\overline{d_H}$; [0.003, 1] for -$w_{min}$; [0.03, 1] for $w_{max}$; [0.003, 3] for $d_s$. For setting their random values, the log-uniform distribution was used. The population size was 300; the elitism level was 0.1; the mutation probability per chromosome was 0.5. The optimization criterion was $R$ (1). It was measured for the last 600 sec of 2000 sec record of ping-pong game where the racket moved chaotically. The total number of rewards was 951. The genetic algorithm terminated when 3 successive generations showed no $R$ increase.

## 3    Results and Discussion

The best result achieved during our experiments occurred in the 17[th] generation of genetic algorithm. The $R$ value reached was 0.553. The optimum parameter values found: $\overline{d_H} = 0.056$; $w_{min}$; = -0.017 (setting it to zero only slightly decreases $R$); $w_{max} = 0.48$ (at least 3 input spikes are required for firing); $d_s. = 0.23$.

Considering the inherent fuzziness in the relationship between the current world state and soon obtaining reward due to the discrete world description and chaotic racket movement, this $R$ value appears satisfactory. To objectively evaluate the performance, we decided to compare our learning neuron with traditional machine learning methods. To ensure a fair comparison, we selected two machine learning algorithms of completely different nature: decision tree and convolutional neural network. All the algorithms were trained on the same binary signal data from the input nodes, with each emulation step serving as a learning example. The target value was binary, labeling target periods. The machine learning algorithms were applied on the same data as our learning neuron (the first 1 400 000 steps) and created scoring models that returned the probability that the current step belongs to the target period. The steps when the score returned exceeds a certain threshold (determined through optimization) were considered analogously to neuron firing. The prediction periods in the last 600 000 simulation steps were determined using the rule described at the beginning of Section 2 and the $R$ value was calculated. The optimum threshold score value was found from the requirement that the $R$ value for the first 1 400 000 steps should be maximum. The decision tree algorithm used the information gain split criterion. The network included 2 convolutional ReLU layers.

The maximum $R$ value obtained by decision tree is 0.742, convolutional network gives $R$ equal to 0.731. The observed proximity of results shown by very different methods proves validity of our approach to determination of theoretical limit for $R$ value in our problem. Thus, the estimation of this limit equal to 0.75 seems to be realistic.

Although the result achieved by our neuron (74% of the theoretical maximum) might be considered modest, we regard our model as successful. Indeed, our neuron is very simple. Being considered as a predictive model, it contains only 133 degrees of freedom. In contrast, the decision tree model includes 51 levels and 403 non-terminal nodes. In essence, the function of our neuron is similar to the conjunction of logical values corresponding to a few of its strongest synapses (see below). Coincidence of these factors is treated as a cause of the target event. However, it is evident that obtaining reward in our example may be preceded by several significantly different conditions. Therefore, it is rather a disjunction of several conjunctions. We propose that a network of these described neurons could yield much more precise reward predictions, and we will explore the potential structure of such a network below.

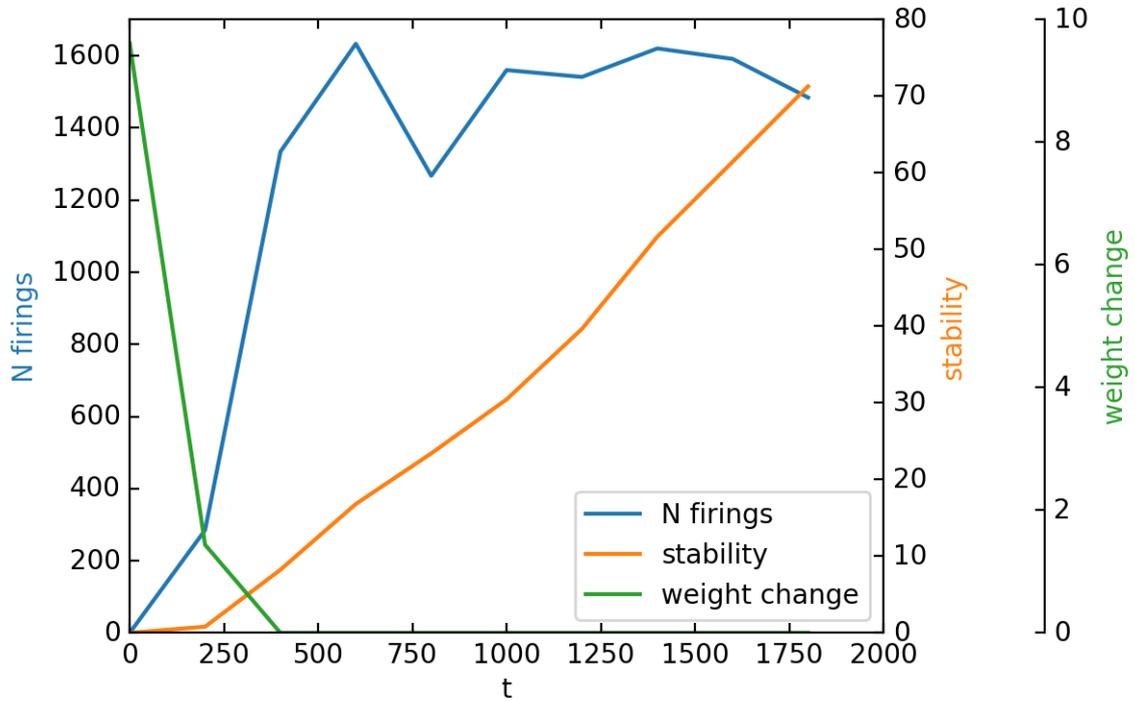

Fig. 1. Dynamics of firing frequency, stability and weight changes of the learning neuron.

Let's examine the learning process and its outcomes. The dynamics of the neuron activity, its stability and the total weight change (the sum of absolute values of the individual synaptic weight changes) are represented on Fig. 1. Since the original weights of all plastic synapses were 0, only the dopamine plasticity worked at first. During the first 100 seconds the neuron did not fire. After 250 seconds, its firing frequency stabilized. Due to the weight stabilization mechanism (notice that the neuron's stability grows almost linearly), the synaptic weights did not change after 600 seconds. In reality, the learning process took 600 seconds instead of the planned 1400 seconds.

The learning results are presented on Fig. 2 depicting values of synaptic resources of the learning neuron at the 2000$^{th}$ second. The leftmost plot corresponds to 30 input nodes coding the ball X coordinate. The vertical axis of all plots except the rightmost one displays synaptic resource value. The second plot corresponds to 30 input nodes coding the Y coordinate of the ball (the blue line) and the racket (the orange line). The next two plots represent 9+9 input nodes coding the horizontal and vertical components of the ball velocity. The rightmost plot shows the color-coded values of synaptic resources of the 25 input nodes indicating the location of the ball within a 5x5 grid that moves with the racket. The distribution of synaptic resource values in these plots appears reasonable and in line with expectations.

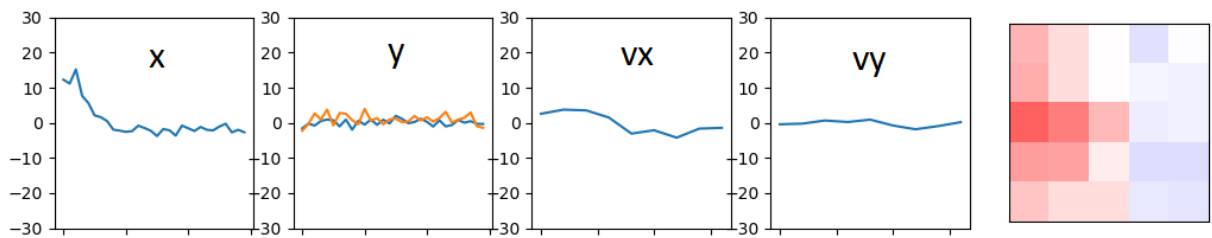

Fig. 2. Synaptic resources of the trained neuron.

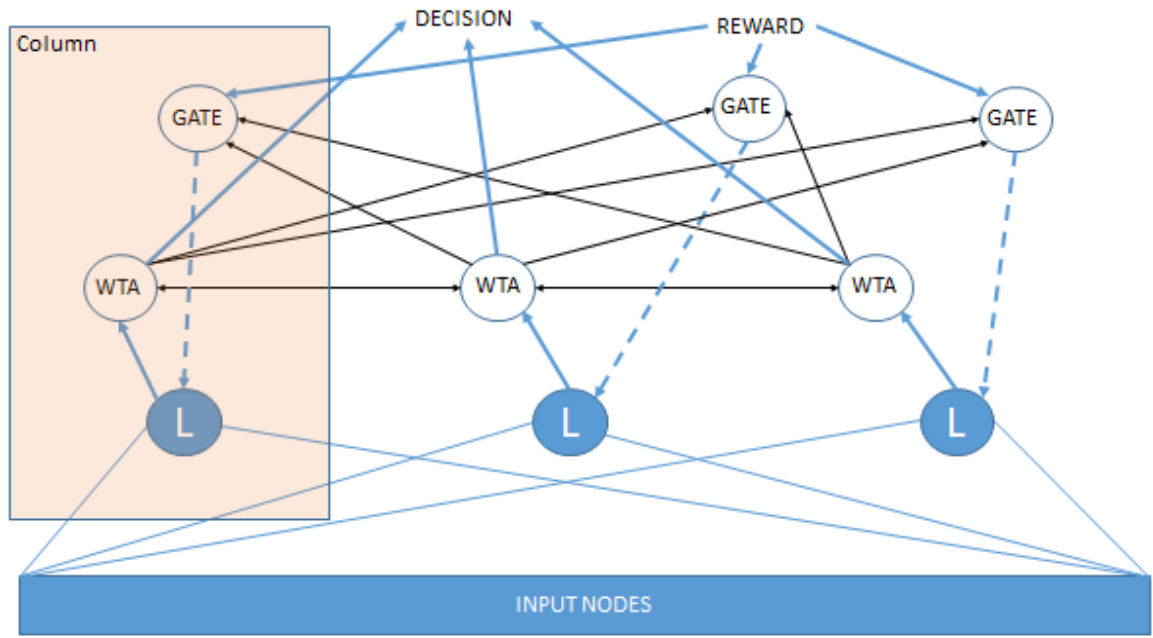

Fig. 3. The possible structure including several neurons recognizing causal links for prediction of the target event which may be a consequence of several significantly different causes. The blue arrows are excitatory connections, the black arrows are blocking, the dashed arrows are reward.

In conclusion, we can assert that our neuron is capable of detecting causal relationships between observed events. However, as previously discussed, a single neuron, even when adequately trained, is not an extremely precise predictor of the imminent occurrence of the target event. This is because the target event may be triggered by several significantly different precursors. To achieve more accurate predictions, a network of neurons of this kind is necessary.

We can propose a hypothesis regarding the potential architecture of such a network (see Fig. 3). In this network, the neurons recognizing different causes of the target event (the blue L rings) enter the columnar structure where each column corresponds to one separate cause. These columns engage in competition to recognize events-causes using lateral inhibitory connections between their "winner-takes-all" (WTA) neurons. The firing ("winning") WTA neuron blocks not only the other WTA neurons but also the GATE neurons in the other columns. If one L neuron fires then the other L neurons should not fire because L neurons should recognize different causal links. If some L neuron fires after the winner it will not receive a reward signal because this signal will not pass through the GATE blocked by the winner. The synapses which caused its firing will be suppressed by anti-Hebbian plasticity and this situation will not repeat next time. If the winning neuron fired correctly it will be rewarded since its GATE is not blocked. We believe that this architectural approach has the potential to recognize complex networks of causal relationships, and we plan to explore and test it in our future research.

## 4      Conclusion.

In conclusion, the ability to discern causal relationships within data streams is a fundamental attribute for any learning system engaging with the real world. In this research, we have showcased the realization of this critical function at the level of an individual neuron, facilitated by a novel combination of anti-Hebbian and dopamine plasticity mechanisms. Given the pivotal role of such mechanisms in the context of implementing reinforcement learning within Spiking Neural Networks (SNNs), we tested our model on a simplified RL problem, namely, the ATARI ping-pong computer game. Our findings, including an estimation of the theoretical upper limit for prediction accuracy in this task, underscore the efficiency of the proposed neuron model in identifying causal links.

Furthermore, we introduced an SNN architecture featuring neurons of the aforementioned type, with the capacity to infer networks of causal relationships directly from raw data. In the forthcoming research, we aim to rigorously test and fine-tune this SNN framework. Additionally, our ongoing investigations will focus on augmenting the capabilities of this SNN to capture temporal aspects of causal relationships, shifting from the question "What are the potential consequences of given events?" to the more nuanced query of "When are these consequences likely to materialize?" This research represents a significant step toward enhancing the capacity of artificial neural networks to model and understand complex causal dynamics in real-world environments.

# References.